# How vehicles change lanes after encountering crashes: Empirical analysis and modeling

**Kequan Chen[a]**
[a]School of Civil and Environmental Engineering
Nanyang Technological University, Singapore 639798
Email: kequan.chen@ntu.edu.sg

**Yuxuan Wang[b*]**
[b]State Key Laboratory of Internet of Things for Smart City
University of Macau, Macau, China
Email: yxwang@um.edu.mo

**Pan Liu[c*]**
[c]School of Transportation, Southeast University
Southeast University Road #2, Nanjing, China, 211189
Email: liupan@seu.edu.cn

**Victor L. Knoop[d]**
[d]Civil Engineering of Technology, Delft University of Technology
Stevinweg 1, Delft, 2628CN
Email: v.l.knoop@tudelft.nl

**David Z. W. Wang[a]**
[a]School of Civil and Environmental Engineering
Nanyang Technological University, Singapore 639798
Email: wangzhiwei@ntu.edu.sg

**Yu Han[c]**
[c]School of Transportation, Southeast University
Southeast University Road #2, Nanjing, China, 211189
Email: yuhan@seu.edu.cn

* Corresponding authors

## ABSTRACT

When a traffic crash occurs, following vehicles need to change lanes to bypass the obstruction. We define these maneuvers as post-crash lane changes (LCs). In such scenarios, vehicles in the target lane may refuse to yield even after the lane change has already begun, increasing the complexity and crash risk of post-crash LCs. However, the behavioral characteristics and motion patterns of post-crash LCs remain unknown. To address this gap, we construct a post-crash LC dataset by extracting vehicle trajectories from drone videos captured after crashes. Our empirical analysis reveals that, compared to mandatory LCs (MLCs) and discretionary LCs (DLCs), post-crash LCs exhibit longer durations, lower insertion speeds, and higher crash risks. Notably, 79.4% of post-crash LCs involve at least one instance of non-yielding behavior from the new follower, compared to 21.7% for DLCs and 28.6% for MLCs. Building on these findings, we develop a novel trajectory prediction framework for post-crash LCs. At its core is a graph-based attention module that explicitly models yielding behavior as an auxiliary interaction-aware task. This module is designed to guide both a conditional variational autoencoder and a Transformer-based decoder to predict the lane changer's trajectory. By incorporating the interaction-aware module, our model outperforms existing baselines in trajectory prediction performance by more than 10% in both average displacement error and final displacement error across different prediction horizons. Moreover, our model provides more reliable crash risk analysis by reducing false crash rates and improving conflict prediction accuracy. Finally, we validate the model's transferability using additional post-crash LC datasets collected from different sites.

## 1. Introduction

Lane-changing (LC) is a fundamental and critical driving task on expressways, as it involves complex interactions among multiple vehicles. Numerous studies have shown that these interactions can lead to several traffic problems, such as traffic crashes (Jamal et al., 2019), deceleration waves (Zheng et al., 2011), and capacity drops (Laval and Daganzo, 2006). Thus, the analysis and modeling of LC maneuvers have long been a central focus in transportation research, offering valuable insights for improving road safety and traffic flow efficiency (Zheng, 2014).

Traditionally, LC maneuvers have been classified into distinct types due to significant behavioral differences. The most common classification is based on the driver's decision-making process, dividing LCs into mandatory lane-changing (MLC) and discretionary lane-changing (DLC) (Pan et al., 2016; Zhang et al., 2024). More recent studies have introduced finer categorizations based on the impacts of LC maneuvers, such as forced LC (Hidas, 2002), failed LC (Ali et al., 2020), cut-in LC (Wang et al., 2019), and negative gap forced LC (Chen et al., 2024). However, most existing studies focus on LC maneuvers under normal traffic conditions, while LCs induced by disruptive events such as traffic crashes remain largely overlooked. When a traffic crash occurs in a specific lane, vehicles traveling in that lane are forced to change lanes when they reach the crash site. We define such LC maneuvers as post-crash LCs. To better illustrate post-crash LCs, **Figure 1** presents two representative cases, each showing a vehicle that initiates a lane change after encountering a crash ahead.

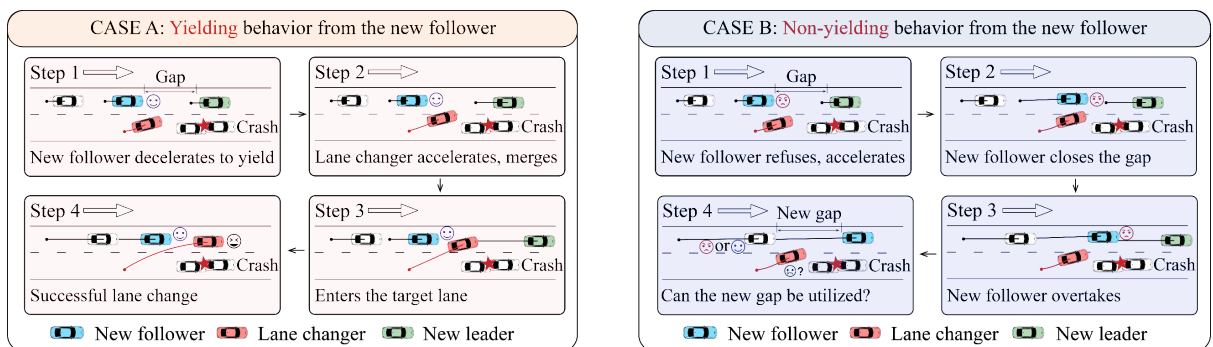

Figure 1 Two cases of lane-changing maneuvers when the lane changer encounters a crash

The LC maneuver in case A in **Figure 1** aligns with natural driving intuition. After detecting a crash ahead, the following vehicle starts to change lanes once it identifies an acceptable gap in the target lane. Meanwhile, the follower in the target lane (new follower) yields to the lane changer, allowing the LC maneuver to be completed. However, it is also common for the new follower to refuse to yield even after the lane changer has commenced. In such cases, the new follower may accelerate and reduce the gap, as shown in steps 1 and 2 of case B in **Figure 1**. Consequently, the lane changer is forced to proceed at a low speed while waiting for another opportunity to merge. Under congested traffic conditions, such non-yielding interactions may occur repeatedly before the lane changer can successfully enter the target lane.

To the best of our knowledge, no empirical studies have examined post-crash LC maneuvers at the individual vehicle level. Consequently, it remains unclear whether real-world post-crash LCs predominantly follow the behavioral pattern of case A or case B. In case B, negative gaps often arise when the new follower overtakes the lane changer. Similar but not identical LC maneuvers, known as negative gap forced LCs (NGFLC), have been documented by Chen et al. (2024), who reported that NGFLCs significantly increase crash risk and reduce traffic flow efficiency. Therefore, ignoring the analysis of post-crash LCs may lead to misinterpretation of how crashes affect downstream traffic flow. A comprehensive understanding of the behavioral features of post-crash LCs and their differences from normal LC maneuvers offers valuable insights. However, such analyses are still absent from the current literature.

Accurately predicting the lane changer's trajectory during post-crash LC maneuvers is also important for various applications, such as traffic safety analysis and post-crash traffic management. Although trajectory prediction for LC maneuvers has been widely studied, most existing models assume that the LC maneuver is continuous and not interrupted by the new follower. However, in post-crash scenarios, the lane changer's trajectory is often affected and even stopped by the non-yielding behavior of successive new followers. Under these conditions, the LC vehicle continuously adjusts its lateral and longitudinal positions while searching for new acceptable gaps throughout the LC maneuver, which further increases the complexity of trajectory prediction. Given that existing models do not account for the characteristics of post-crash LCs, it becomes critical to question whether they remain applicable to such scenarios.

To address the above issues, this study is the first to investigate how lane changers interact with surrounding vehicles during post-crash LC maneuvers, and to model the lane changer's trajectory accordingly. To support this research, we conduct year-long drone video recordings at a merging section in Nanjing, China, capturing 50 crash events. The LC maneuvers following these crashes form the basis for both our empirical analysis and model development. As such, the primary contributions of this study are summarized as follows:

1. We conduct a comprehensive analysis of post-crash LC maneuvers, focusing on key behavioral features including yielding behavior, LC duration, LC speed, speed differentials, and crash risk. These results are further compared with those of MLCs and DLCs. We highlight the unique behavioral patterns of post-crash LCs, in particular the frequent occurrence of non-yielding behavior by the new follower.

2. We propose a novel deep learning framework for predicting the future trajectory of the lane changer during the post-crash LC maneuver. The yielding behavior of the new follower is explicitly modeled as an auxiliary task to guide the prediction process. This design enhances both interpretability and predictive performance, as validated by ablation and comparative experiments. It also leads to more accurate crash risk analysis.

3. We collect additional data from different sites and cities to evaluate the transferability of the proposed model. Results demonstrate that our model maintains robust predictive accuracy across a wide range of post-crash LCs.

The rest of this paper is organized as follows. Section 2 provides a comprehensive literature review. Section 3 presents empirical data and a preliminary analysis of post-crash LCs. Section 4 presents methodologies for predicting the lane changer's future trajectory. Section 5 designs experiments. Section 6 evaluates the performance of our proposed model. Section 7 concludes our work and offers future research directions.

**2. Related work**

Given the limited research available on post-crash LC maneuvers, we first review existing studies on other LC types. These works offer useful references for understanding the unique features of post-crash LCs. We next review trajectory prediction models for LC maneuvers.

## 2.1 Different types of lane-changing maneuvers

In traditional traffic flow theory, MLC and DLC have been the primary focus of LC behavior research. MLC refers to a situation where the driver is required to change lanes to comply with a specific route, while DLC occurs when the driver changes lanes to obtain a better driving environment (Yang and Koutsopoulos, 1996; Zheng, 2014). Numerous studies have examined the features of MLCs and DLCs using empirical data, including gap acceptance behavior (Vechione et al., 2018; Yang et al., 2019), duration of LC (Li et al., 2022; Wang et al., 2014), merging speed (Daamen et al., 2010; van Beinum et al., 2018), and crash risk (Ali et al., 2022b, 2019). Findings from these studies consistently demonstrate significant differences in behavioral features between MLCs and DLCs. Then, researchers conducted in-depth investigations on each type individually. For example, Gong and Du (2016) considered the impact of MLCs on determining the optimal locations for advanced lane-change warning systems. Ali et al. (2018) explored the effects of connected vehicle environments on MLCs using a driving simulator. They found that connected conditions lead to safer but slower MLCs, as evidenced by larger gap acceptance, lower crash risk, and longer duration. For DLC, Balal et al. (2016) developed a Fuzzy Inference System (FIS) to model the binary decision-making process of whether and when to initiate a DLC maneuver on freeways. Zhang et al. (2023) emphasized the role of individual driving styles in DLCs. Then, they proposed a personalized decision-making framework that learns from the ego vehicle's behavioral styles.

With the advancement of data collection technologies, the classification of LC maneuvers has become more detailed. One such category is failed LCs, where a driver starts an LC attempt but eventually aborts the maneuver and returns to the original lane (Lee, 1996). In recent years, this type of LC maneuver has gained growing research attention, with studies focusing on its impact (Ali et al., 2022a), predictive classification methods (Shi and Liu, 2019; Sun et al., 2014), crash risk (Ali et al., 2020), and influencing factors (Xing et al., 2025). Another notable category is the cut-in LC, which occurs when the lane changer merges into the target lane with a very short longitudinal gap from the new follower (Sultan et al., 2002). Due to its implications for traffic safety and roadway capacity, this behavior has also attracted significant attention. For example, Wang et al. (2019) analyzed 5,608 cut-in LC events from the Shanghai Naturalistic Driving Study and found that nearly half of the drivers did not use turn signals and that cut-in LCs were associated with higher crash risks compared to regular LC maneuvers. Hu et al. (2024) developed an eco-driving strategy for connected autonomous vehicles (CAVs) that accounts for cut-in LC by manually driven vehicles. When the gap between the lane changer and the new follower turns negative, a cut-in LC may escalate into a negative gap forced LC (NGFLC). According to Chen et al. (2024), NGFLCs form the basis of sideswipe crashes and contribute to elevated crash risk and capacity reduction in the target lane. More recently, Wang et al. (2024) challenged the traditional assumption that MLCs must always be executed. By quantifying the costs associated with missed freeway exits through analytical modeling, they proposed a new classification termed expedient lane change (ELC), reflecting cases where bypassing an MLC may be a rational choice.

## 2.2 Lane-changing trajectory prediction models

Over the past few decades, numerous studies have been conducted on predicting the trajectory of lane changers during LC maneuvers. These prediction models can be divided into two groups: mathematical function-based models and data-driven models. In complex traffic environments, mathematical function-based models struggle to capture the nonlinear interactions between the lane changer and multiple surrounding vehicles. In contrast, data-driven models, leveraging advanced pattern recognition and learning capabilities, have shown superior performance in trajectory prediction (Chen et al., 2023; Dou et al., 2016). As a result, data-driven approaches have become the predominant methodology for LC trajectory prediction in recent years. For example, Xie et al. (2019) proposed one of the earliest deep learning frameworks for modeling the entire LC maneuver, employing a Deep Belief Network (DBN) for LC decisions and a long short-term memory (LSTM) network for LC execution. Xue et al. (2022) developed an integrated framework that incorporates traffic conditions and vehicle types to improve adaptability across environments. Lu et al. (2025) proposed a Transformer-based transfer learning framework that addresses the limitations of the fixed neighbor encoding and the underutilization of trajectory data. Li et al. (2025) further advanced this field by introducing an LSTM framework that integrates individual driving styles to improve trajectory prediction performance for autonomous vehicle applications.

Indeed, there also have been several studies that, although not specifically designed for LCs, have demonstrated their models' applicability to lane changers' trajectory prediction during evaluation. For example, Liu et al. (2022) developed a conditional variational autoencoder (CVAE) based framework augmented with a driving risk map. This model accounts for motion uncertainty, intention ambiguity, and vehicle-environment interactions, leading to improved trajectory prediction accuracy. Qian et al. (2025) integrated a BiLSTM-based trajectory generation, Conv1D-based intention recognition, and consistency regularization constraints to predict vehicle trajectory. Liao et al. (2024) introduced HLTP, a human-inspired trajectory prediction model that employs a teacher-student knowledge distillation framework to refine trajectory learning. A more comprehensive overview of trajectory prediction research can be found in the surveys by Huang et al. (2022), Ding and Zhao (2023), and Rowan et al. (2025).

The methods mentioned above improve trajectory prediction accuracy by integrating various contextual features or optimizing network structures. However, most of these approaches model the interactions between the lane changer and surrounding vehicles in an implicit manner. This often limits interpretability and makes it challenging to understand how specific interactions influence the predicted trajectories. Recently, some studies have shown that explicitly encoding key behavioral cues can significantly enhance both the interpretability and performance of prediction models (Geng et al., 2023; Ivanovic et al., 2021; Zhang et al., 2023; Zhou et al., 2023).

Inspired by these observations, we revisit two representative post-crash LC cases, as illustrated in **Figure 1**. It is evident that the lane changer's movement patterns vary significantly based on whether the new follower yields. Motivated by this observation, this study aims to explicitly model and predict the yielding behavior of the new follower and incorporate it into a deep learning framework to predict the lane changer's trajectory. This design is expected to improve both the interpretability and accuracy of the prediction model.

## 3. Data

### 3.1 Study site

Due to the rarity and randomness of crashes, collecting post-crash LC data is particularly challenging. To address this, we conduct one year of on-site drone video recordings at a busy merging section in Nanjing. The location and layout of the study site are shown in **Figure 2**. This site is selected for its high crash frequency, which increases the likelihood of observing post-crash LCs through continuous recording. Although this data collection process is labor-intensive, it enables the acquisition of high-quality microscopic traffic data. Specifically, drone video is recorded during the morning peak hours (7:30 AM to 9:30 AM) on each working day from October 1, 2021 to October 1, 2022. The drones operate at an altitude of 300 meters, covering about 400 meters of roadway. They are equipped with high-resolution cameras that record 4K video at 30 frames per second.

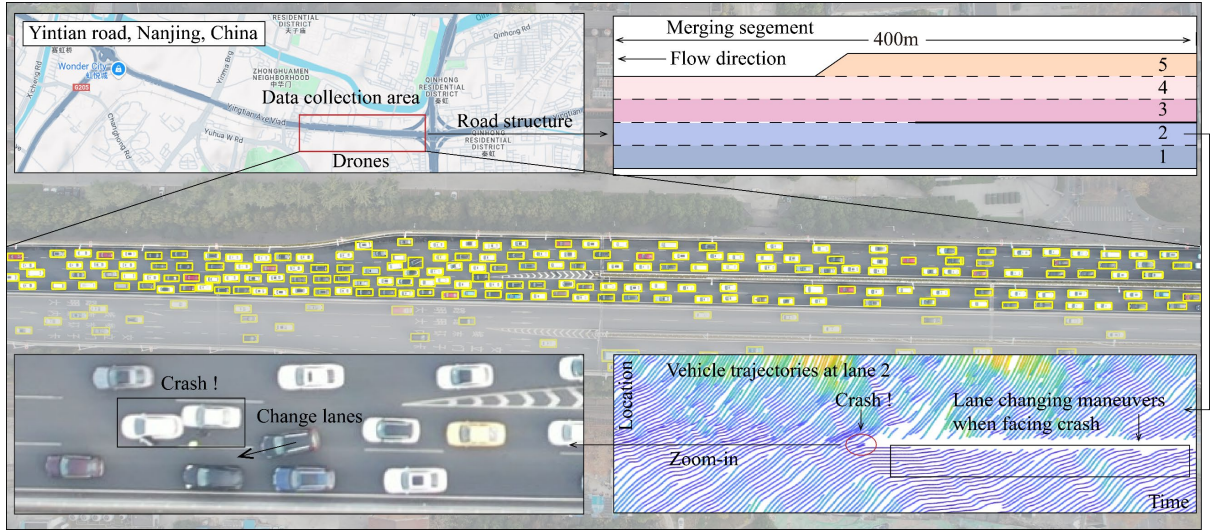

Figure 2 Study site and vehicle trajectories

During the data collection period, we obtain over 200 hours of effective drone videos. In total, we record 50 traffic crashes. For each crash-involved drone video, vehicle trajectories are extracted using a high-precision trajectory extraction framework. In this framework, the raw footage is first stabilized to remove camera motion caused by wind and manual operation. A customized vehicle detection module achieves an accuracy exceeding 99%, with a mean error of 0.04 m between the detected and true vehicle centers. Based on these detections, a kinematics-driven tracking algorithm reconstructs continuous trajectories for all vehicles. Finally, a smoothing procedure is applied to remove high-frequency noise, yielding stable and physically consistent motion profiles suitable for subsequent analysis. Interested readers may refer to our earlier publications for details on this trajectory extraction framework (Chen et al., 2023; Chen et al., 2024; Wang et al., 2024). An example of the extracted trajectories following a crash is shown in the lower part of **Figure 2**. In this example, a rear-end crash occurs in lane 2, and following vehicles in this lane have to perform LC maneuvers as they approach the crash site. These post-crash LC maneuvers, highlighted by the black rectangular region in the trajectory plot shown in **Figure 2**, form the core dataset for our analysis. It is worth noting that some crashes occurred near the lane boundaries. In such cases, following vehicles are able to bypass the crashed vehicle at low speeds without performing an LC. These cases are excluded from our analysis, as they do not constitute complete LC maneuvers. After this filtering process, a total of 1,374 post-crash LC samples are identified for subsequent analysis and model development.

### 3.2 Preliminary Analysis

To facilitate intuitive understanding, we illustrate a representative post-crash LC maneuver in **Figure 3**. This plot depicts the lateral position of the lane changer over time, with speed encoded by color. In the following, we analyze interactions between the lane changer and the new follower at several critical time points.

Time $t_1$ marks the starting time of the LC maneuver. We identify this moment using the Mexican-hat wavelet transform on the lateral position of the lane changer, as suggested by previous studies (Ali et al., 2022a; Chen et al., 2021; Zheng et al., 2011). Although a gap (gap 1 in **Figure 3**) is present in the target lane at $t_1$, the new follower (NF) does not yield to the lane changer. At $t_2$, the NF closes this gap and overtakes the lane changer. At the same time, a new gap (gap 2 in **Figure 3**) between the NF and its follower (denoted as NF2) emerges. However, NF2 also refuses to yield and overtakes the lane changer by $t_3$, which subsequently leads to the formation of a third gap (gap 3 in **Figure 3**) between NF2 and its following vehicle (NF3). Ultimately, NF3 yields to the lane changer. The lane changer uses gap 3 to complete the LC maneuver from $t_3$ to $t_6$. Time $t_6$ is the ending time of the LC maneuver, which is also identified using the wavelet transform approach.

Based on the above analysis, the entire post-crash LC maneuver can be segmented according to the yielding behavior of the new follower. Thus, the yielding behavior of the new follower is identified based on its kinematic response to the lane changer. Specifically, a new follower is considered to exhibit non-yielding behavior if it accelerates or advances in a manner that closes the available gap and prevents the lane changer from merging. Conversely, yielding behavior is identified when the new follower maintains or enlarges the gap, allowing the lane changer to complete the merge using that gap. From $t_1$ to $t_3$, the lane changer encounters consecutive non-yielding responses, while from $t_3$ to $t_6$, the new follower exhibits yielding behavior toward the lane changer.

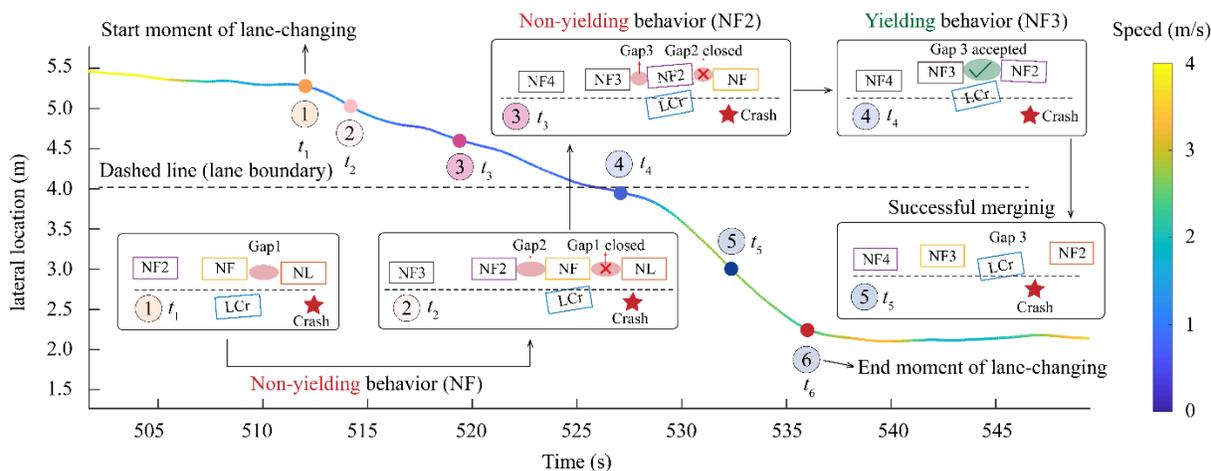

Figure 3 An example of a post-crash lane-changing maneuver

Here, we compare the behavioral characteristics of post-crash LCs with those of DLCs and MLCs. DLC samples are selected from lanes 1, 2, and 3 at the study site. MLC samples are drawn from lane changes occurring in the merging lane (lane 5), where vehicles are required to change lanes due to geometric constraints. To ensure a fair comparison, the number of DLC and MLC samples is matched to that of post-crash LC samples.

We first investigate the number of rejected gaps during each LC maneuver. Here, a rejected gap is defined as a gap in the target lane that becomes available during a post-crash LC maneuver but is not used for the final merge. The comparison results are shown in **Figure 4(a)**. From this figure, it is observed that around 79.4% of post-crash LCs involve at least one rejected gap, while only 21.7% of DLC samples and 28.6% of MLC samples exhibit similar behavior. This finding suggests that the gap rejection behavior illustrated in **Figure 3** is a typical feature of post-crash LCs, which distinguishes them from regular LC types. As a consequence of the higher frequency of gap rejections, post-crash LCs exhibit a significantly longer average duration (18.49s) compared to MLCs (12.41 s) and DLCs (10.87 s), as shown in **Figure 4(b)**. Here, LC duration is defined as the time interval between the start and end moments of the LC maneuver.

The initial speed of the lane changer across different LC types is summarized in **Figure 4(c)**. The mean initial speed in post-crash LCs (1.33 m/s) is notably lower than that in DLCs (3.56 m/s) and MLCs (4.65 m/s). This disparity is primarily due to the fact that post-crash LCs occur under congested conditions, where the crash obstructs traffic and forces vehicles to decelerate before initiating an LC maneuver. In contrast, vehicles in the target lane are less affected by the crash, resulting in a larger speed difference between the lane changer and the new follower, as shown in **Figure 4(d)**. This indicates that new followers are moving much faster than the lane changer. Such a speed mismatch complicates the execution of the LC maneuver, often requiring multiple attempts to find an acceptable gap before a successful merge. For DLC and MLC, the LC maneuver usually occurs after the lane changer finds a suitable gap. Consequently, the initial speed of the lane changer tends to be higher, and the speed difference with the new follower is much smaller. These results are shown in **Figure 4(c)** and **(d)**. This aligns with the general understanding of typical LC behavior.

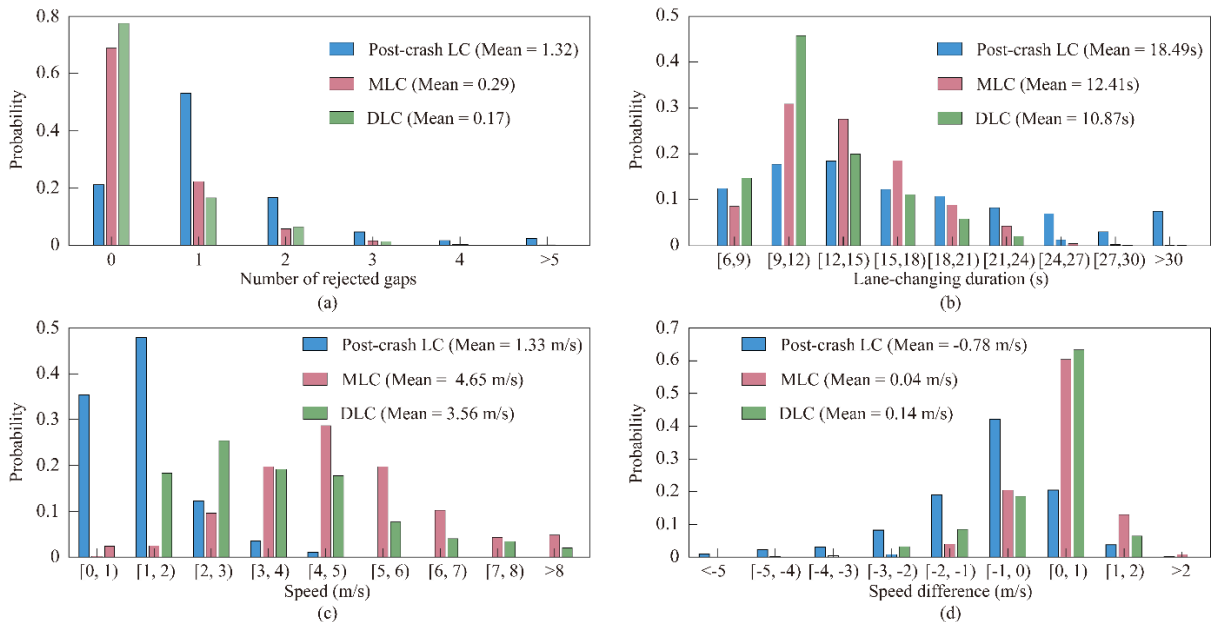

Figure 4 Comparison of lane-changing characteristics among post-crash LCs, MLCs, and DLCs: (a) number of rejected gaps; (b) lane-changing duration; (c) initial speed of the lane changer; and (d) Speed difference between the lane changer and the new follower

For each post-crash LC sample with at least one rejected gap, we compute the proportion of time during which the new follower exhibits non-yielding behavior, relative to the total duration of the LC maneuver. The statistical results are presented in **Figure 5(a)**. For comparison, the same proportion is calculated for DLCs and MLCs that also exhibit rejected gaps. From **Figure 5(a)**, we notice that post-crash LCs have a significantly higher proportion of non-yielding time. On average, non-yielding behavior accounts for nearly 50% of the total LC duration in post-crash LCs. In contrast, this proportion is much lower in DLCs and MLCs, measuring 17.4% and 21.3%, respectively. The Kruskal–Wallis test reveals a statistically significant difference across the three LC types ($p < 0.001$). Post-hoc Dunn's tests show that post-crash LCs have a significantly higher proportion of non-yielding time than both MLCs and DLCs.

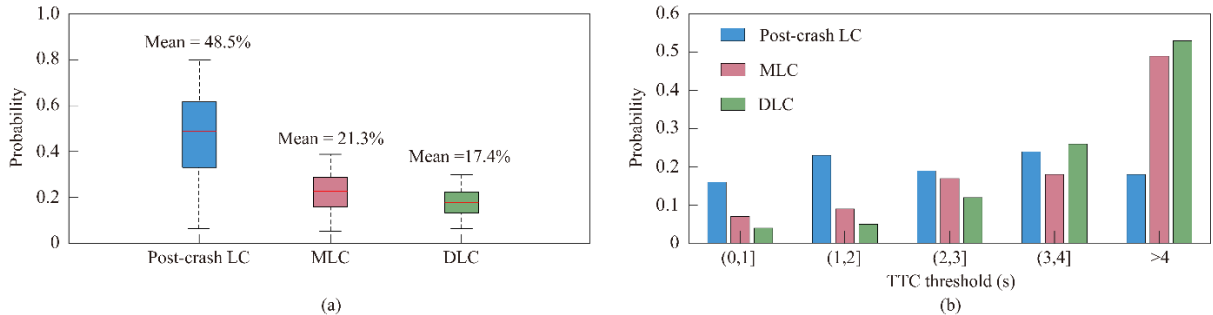

Figure 5 (a) Proportion of non-yielding behavior duration; (b) Distribution of conflict probability under different TTC thresholds

Finally, we evaluate the crash risk associated with different LC types using a two-dimensional Time-to-Collision (TTC) metric. Unlike the conventional one-dimensional TTC, which considers only longitudinal interactions, the two-dimensional TTC accounts for both longitudinal and lateral vehicle motions and is therefore more suitable for characterizing interactive lane-changing scenarios. Previous studies have shown that this metric provides a more accurate representation of collision risk during lane-changing maneuvers (Chen et al., 2024; Zhang et al., 2023). The two-dimensional TTC is computed under the assumption that both vehicles maintain constant velocities and headings, and is defined as the time until the projected rectangular footprints of the two vehicles first overlap. For each LC event, we calculate the TTC between the lane changer and surrounding vehicles at each time step, with the minimum value adopted to represent the crash risk of that LC maneuver. The proportion of LC samples at different TTC thresholds is summarized in **Figure 5(b)**. Results show that 41% of post-crash LCs have a TTC below 2s, which is a commonly used threshold to distinguish between risky and safe interactions. In comparison, only 16% of MLCs and 9% of DLCs fall below this threshold. When applying a more stringent threshold of 1 s to identify high-risk conflicts, 16% of post-crash LCs meet this criterion, significantly higher than the 6% of MLCs and 4% of DLCs. These results highlight that post-crash LCs face a higher crash risk than regular LCs.

## 4. Methodologies

The interaction between the lane changer and the new follower makes post-crash LC maneuvers distinct from traditional LC maneuvers in terms of several key behavioral features. In this section, we present a novel framework to predict the lane changer's trajectory by explicitly modeling the interaction behavior between the lane changer and the new follower. The overall architecture of our proposed method is illustrated in **Figure 6**. We begin by introducing the model's inputs and outputs in **Section 4.1**. The construction and integration of the three key modules are described in the following sections: a Conditional Variational Autoencoder (CVAE) for capturing uncertainty in trajectory prediction (**Section 4.2**); a Graph Attention Network (GAT) for modeling interactions as an auxiliary task (**Section 4.3**); a Transformer-based decoder for refining coarse trajectory predictions (**Section 4.4**). The entire framework is trained in an end-to-end fashion, with the loss function design detailed in **Section 4.5**. As our model integrates the **C**VAE, **I**nteraction-aware module, and **T**ransformer-based decoder, we refer to it as **CIT**.

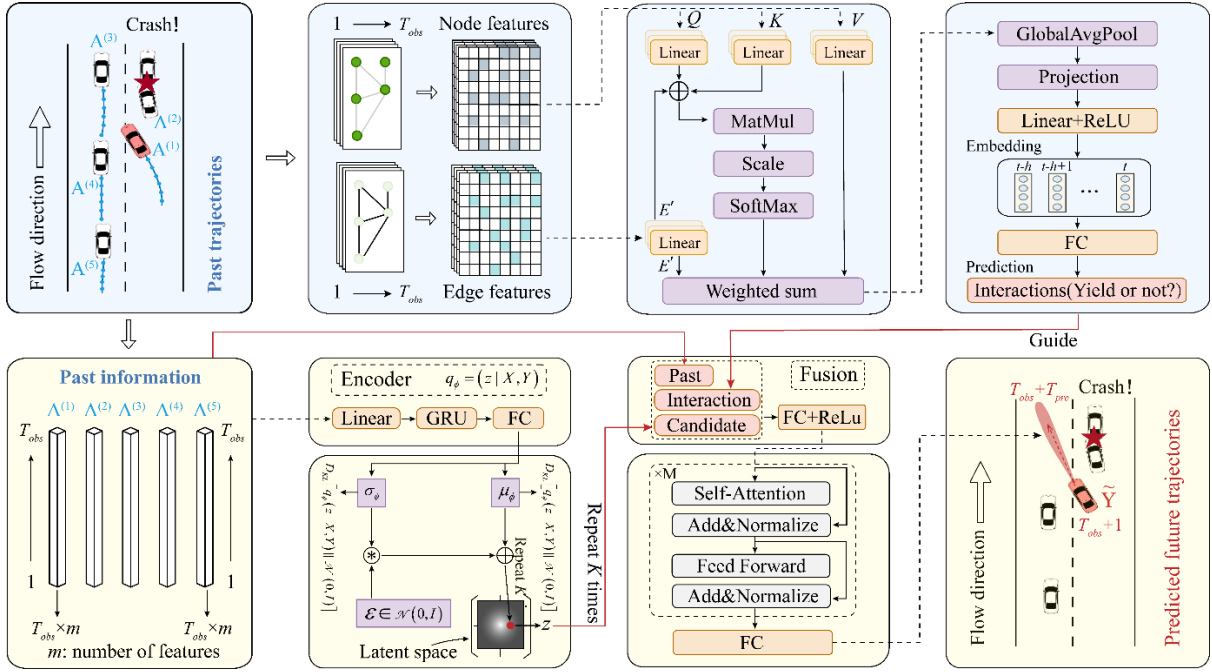

Figure 6 Overall architecture of CIT

## 4.1 Problem Formulation

For each time step $t$, the state of vehicle $p$ is defined as $A_p^{(t)} \in \mathbb{R}^5$, which includes the lateral position, longitudinal position, speed, acceleration, and steering angle. In a complete post-crash LC maneuver, we consider five related vehicles: the lane changer ($p$=0), the crashed vehicle ($p$=1), the new leader ($p$=2), the new follower ($p$=3), the follower after the new follower ($p$=4). Over an observation time horizon of $T_{obs}$, we define $X = \{A_0^{(t)}, A_1^{(t)}, A_2^{(t)}, A_3^{(t)}, A_4^{(t)}\}_{t=1}^{T_{obs}} \in \mathbb{R}^{T_{obs} \times 5 \times 4}$ as the model's input. Following the analysis in **Section 3.2**, we label the ground truth of the new follower's yielding behavior based on the final gap into which the lane changer merges. If the lane changer rejects one or more gaps before accepting the final gap, the new follower is labeled as non-yielding until the final gap becomes available. Once the final gap appears, the new follower's behavior becomes yielding. In cases where the first gap is also the final gap, the entire post-crash LC maneuver is considered a yielding instance. Based on this identification scheme, we define $B = \{\delta^{(t)}\}_{t=T_{obs}+1}^{T_{obs}+T_{pre}} \in \mathbb{R}^{T_{pre} \times 1}$ to denote the ground truth interaction behavior over the prediction interval $[T_{obs}+1, T_{obs}+T_{pre}]$, where $\delta^{(t)} = 1$ indicates yielding behavior and $\delta^{(t)} = 0$ indicates non-yielding behavior. Here, $T_{pre}$ represents the future prediction horizon. The output of our model is defined as $\hat{Y} = \{\hat{Y}^{(t)}\}_{t=T_{obs}+1}^{T_{obs}+T_{pre}} \in \mathbb{R}^{T_{pre} \times 2}$, representing the predicted trajectory of the lane changer over the prediction horizon $T_{pre}$, where $\hat{Y}^{(t)} = \{\hat{x}^{(t)}, \hat{y}^{(t)}\}$ includes the longitudinal and lateral positions at future time step $t$.

Based on the abovementioned identification, we develop a framework described in the following sections that leverages the predicted interaction sequence $\hat{B}$ to guide trajectory prediction $\hat{Y}$ from the historical vehicles' states $X$, expressed as $\hat{Y} = F(X; \hat{B})$.

**4.2 Conditional Variational Autoencoder**

When a driver notices that the preceding vehicle has crashed, they may exhibit multiple plausible trajectories to the target lane due to the stochastic nature of driving behavior. An effective model should capture this uncertainty within reasonable bounds. To this end, we adopt the Conditional Variational Autoencoder (CVAE), a widely used deep generative model (Lee et al., 2017). A standard CVAE architecture comprises three sub-networks, i.e., a recognition network $q_\phi(z|Y, X)$ where $z$ is the latent variable, a prior network $p_\phi(z|X)$, and a generation network $p_\phi(\hat{Y}|X, z)$. To implement the prior network, we first compress the input sequence $X$ into a feature representation using a two-stage encoder. At each time step $t$ ($1 \leq t \leq T_{obs}$), the historical state $X^{(t)}$ is first passed through a multi-layer perceptron (MLP) to obtain a latent embedding. This embedding, together with the hidden state from the previous time step $h^{(t-1)}$, is then fed into a gated recurrent unit (GRU), producing the updated hidden state $h^{(t)} \in \mathbb{R}^{d_h}$:

$$h^{(t)} = GRU\left(MLP\left(X^{(t)}; \Theta_{MLP}\right), h^{(t-1)}; \Theta_{GRU}\right) \tag{1}$$

where the initial hidden state $h^{(0)}$ equals zero, and $\Theta_{MLP}$ and $\Theta_{GRU}$ are the trainable parameters of MLP and GRU, respectively.

After applying **Equation (1)** from $t = 1$ to $T_{obs}$, the input sequence $X$ is encoded into the final hidden representation $h^{(T_{obs})}$.

The latent variable $z$ is assumed to follow a Gaussian distribution $\mathcal{N}(\mu, \sigma^2)$, where $\mu$ and $\sigma^2$ represent the mean and variance, respectively. To estimate these parameters, the final representation $h^{(T_{obs})}$ is passed through two separate fully connected layers:

$$\mu = W_\mu \cdot h^{(T_{obs})} + b_\mu \tag{2}$$

$$\log \sigma^2 = W_\sigma \cdot h^{(T_{obs})} + b_\sigma \tag{3}$$

where $W_\mu$ and $b_\mu$ are the weight matrix and bias vector used to map $h^{(T_{obs})}$ to $\mu$. Likewise, $W_\sigma$ and $b_\sigma$ are the weight matrix and bias vector for producing the $\log\sigma^2$. Predicting $\log\sigma^2$ ensures that the resulting $\sigma^2$ remains positive after exponentiation, and also improves numerical stability during training.

For stochastic inference, we apply the reparameterization trick to obtain stochastic samples $z^{(k)}$ (Kingma and Welling, 2013):

$$z^{(k)} = \mu + \sigma \odot \varepsilon^{(k)} \tag{4}$$

where $\odot$ denotes the element-wise multiplication, $\varepsilon$ is an auxiliary noise variable, and $\varepsilon^{(k)}$ is drawn from a standard normal distribution $\mathcal{N}(\mathbf{0}, \mathbf{I})$.

By repeating **Equation (4)** for $k=1$ to $K$, we obtain a set of latent samples denoted as $C = \{z^{(1)}, z^{(2)}, ..., z^{(K)}\} \in \mathbb{R}^{K \times d_z}$. In conventional CVAE-based prediction frameworks, this latent set is then passed to the generation network to produce diverse candidate trajectories. In our framework, we enhance this process by incorporating the predicted interaction behavior between the lane changer and the new follower. The purpose of the interaction-aware mechanism is to guide the generation network to produce more human-like and realistic post-crash LC trajectories. In the following section, we detail how the interaction behavior is predicted and how it is fused with the latent set $C$ to inform trajectory generation.

### 4.3 Interaction-aware module

The prediction process for the yielding behavior of the new follower is shown in the upper part of **Figure 6**. We use a graph structure to capture how interactions among multiple vehicles influence the new follower's yielding behavior during the post-crash LC. At each observation time step $t \in [1, T_{obs}]$, we define the graph $\mathcal{G} = \{N^{(t)}, E^{(t)}\}_{t=1}^{T_{obs}}$. Here, $N^{(t)} = \{n_i^{(t)} | \forall i \in 1, ..., v\} \in \mathbb{R}^{v \times d_n}$ is the set of $v$ nodes at time $t$. Each node vector $n_i^{(t)} \in \mathbb{R}^{d_n}$ has $d_n$ dimension features, which include instantaneous speed, acceleration, steering angle, lateral position, and longitudinal position. The edge set at time $t$ is represented as $E^{(t)} = \{e_{ij}^{(t)} | \forall i, j \in 1, ..., s; i \neq j\} \in \mathbb{R}^{r \times d_e}$, where $r$ is the number of edges. Each edge vector $e_{ij}^{(t)} \in \mathbb{R}^{d_e}$ has $d_e$ dimension features, which include the relative speed and acceleration, and lateral and longitudinal distance.

To extract interaction-aware representations from the graph $\mathcal{G}$, we first apply a self-attention mechanism to the node feature vectors $n_i^{(t)}$. For each attention head $m$ at time $t$, we use three separate linear transformations to obtain queries $Q_i^{(m),t}$, keys $K_i^{(m),t}$, and values $V_i^{(m),t}$ from $n_i^{(t)}$. This process is defined as:

$$Q_i^{(m),t} = W_Q^{(m)} n_i^{(t)}, \ K_i^{(m),t} = W_k^{(m)} n_i^{(t)}, \ V_i^{(m),t} = W_v^{(m)} n_i^{(t)} \tag{5}$$

where $W_Q^{(m)}$, $W_k^{(m)}$, and $W_v^{(m)}$ are trainable weight matrices.

We integrate edge vectors $e_{ij}^{(t)}$ into the node-level attention mechanism by projecting them through a linear mapping. For head $m$ at time $t$, each edge vector $e_{ij}^{(t)}$ is mapped by a learnable weight $W_M^{(m)}$ to generate an attention bias term $W_{ij}^{(m),t}$. This bias is then added to the raw attention scores computed from the queries and keys, which can be calculated as below:

$$R_{ij}^{(m),t} = \frac{Q_i^{(m),tT} \left( K_i^{(m),t} + M_{ij}^{(m),t} \right)}{\sqrt{d}} \tag{6}$$

Then, the attention coefficient of node $j$ with respect to node $i$ for head $m$ at time $t$ is concatenated as follows:

$$\alpha_{ij}^{(m),t} = \frac{\exp\left(R_{ij}^{(m),t}\right)}{\sum_{k \in \mathcal{N}_i(j)} \exp\left(R_{ik}^{(m),t}\right)} \tag{7}$$

where $\mathcal{N}_i(j)$ is the set of nodes connected to node $i$.

Based on the attention coefficients of edges incident to node $i$, we perform a weighted aggregation of the neighboring node features to compute the updated representation of node $i$ for head $m$ at time $t$. The formula is given as follows:

$$h_i^{(m),t} = softmax\left[\sum_{k \in \mathcal{N}_i(j)} \alpha_{ij}^{(m),t} \left(V_i^{(m),t} + M_{ij}^{(m),t}\right)\right] \tag{8}$$

To combine the outputs of $M$ attention heads, the head-specific embeddings at node $i$ and time $t$ are concatenated as below:

$$h_t^t = h_t^{(1),t} \| ... \| h_t^{(m),t} \tag{9}$$

Global average pooling is applied to compress the node representations into a single graph-level vector $\gamma^t \in \mathbb{R}^{d_\gamma}$, as defined in **Equation (10)**.

$$\gamma^t = \frac{1}{|v|} \sum_{i \in v} h_i^t \tag{10}$$

After pooling, we project the output into a new feature space with a learnable linear mapping.

$$q^t = W_q \gamma^t + b_q \tag{11}$$

where $W_q \in \mathbb{R}^{d_q \times d_\gamma}$ are the weight matrices to map $\gamma^t$ into $\mathbb{R}^{d_q}$, $b_q \in \mathbb{R}^{d_q}$ is the bias.

Then, a linear layer followed by a ReLU activation is used to encode the graph's state at time $t$ in a form that is convenient for prediction.

$$e^t = ReLU\left(W_e q^t + b_e\right) \tag{12}$$

where $W_e \in \mathbb{R}^{d_e \times d_q}$ and $b_e \in \mathbb{R}^{d_e}$ are the weight and bias of the linear network.

To capture dynamics over the observation time window, we concatenate per-step features into a single vector:

$$E = \left[e^{1^T}, e^{2^T}, ..., e^{T_{obs}^T}\right] \tag{13}$$

Finally, we feed this temporal embedding $E \in \mathbb{R}^{T_{obs} \times d_e}$ through a fully connected layer with sigmoid activation to produce the predicted interaction behavior $\hat{B}$ over the future horizon [$T_{obs}+1$, $T_{obs}+T_{pre}$].

$$\hat{B} = Sigmoid(W_o E + b_o) \tag{14}$$

### 4.4 Lane-changing trajectory generation module

From the previous sections, we have three inputs: the predicted interaction behavior $\hat{B}$, the latent set $C$, and the historical trajectory $X$. To let $\hat{B}$ guide trajectory generation conditioned on $C$, we first fuse these features with an MLP:

$$e_{fuse} = f_{fuse}(X, C, \hat{B}) \tag{15}$$

$$f_{fuse}(\cdot) = MLP([.]; \Theta_{MLP}) \tag{16}$$

A Transformer network is then utilized to capture the temporal dependencies in the historical time-series observations. The output $e_{fuse}$ is fed into this decoder, which produces a hidden feature sequence $H_{dec} \in \mathbb{R}^{T_{pre} \times d_{trans}}$.

$$H_{dec} = Transformer(e_{fuse}; \Theta_{trans}) \tag{17}$$

where $\Theta_{trans}$ are the Transformer's trainable parameters, and $d_{trans}$ is the feature dimension of the Transformer network.

Finally, a fully connected predictor $p_\phi$ maps $H_{dec}$ to the future predicted trajectory of the lane changer $\hat{Y} \in \mathbb{R}^{T_{pre} \times 2}$ over the interval $[T_{obs}+1, T_{obs}+T_{pre}]$.

$$\hat{Y} = p_\phi(H_{dec}; \Theta_\phi) = \{(\hat{x}^{(t)}, \hat{y}^{(t)}) \mid \forall t \in T_{obs}+1, T_{obs}+2, .., T_{obs}+T_{pre}\} \tag{18}$$

where $\Theta_\phi$ are the trainable weight matrices of the fully connected predictor $p_\phi$.

### 4.5 Loss functions

Our model is trained end-to-end by back-propagation with three complementary losses. The first is the reconstruction loss $\mathcal{L}_p$. It measures the squared error between the ground truth trajectory and the predicted future trajectory generated by our proposed model. The formula $\mathcal{L}_p$ is as follows:

$$\mathcal{L}_p = \frac{1}{M} \sum_{i=1}^{M} \sum_{t=T_{obs}+1}^{T_{obs}+T_{pre}} \left\| Y_i^t - \hat{Y}_i^t \right\|^2 \tag{19}$$

where $M$ is the number of samples in the batch, $\|.\|$ denotes the Euclidean norm, $Y_i^t$ is the real post-crash LC trajectory at time $t$, and $\hat{Y}_i^t$ is the predicted post-crash LC trajectory at time $t$.

The second loss term is the KL divergence $\mathcal{L}_{KL}$. It regularizes the latent space by encouraging the approximate posterior $q_\phi(z|Y,X)$ to align with the prior distribution $p_\phi(z|X)$. Minimizing $\mathcal{L}_{KL}$ prevents overfitting by limiting the amount of information stored in the latent variable $z$. It also ensures that, at inference time, latent samples drawn from the prior yield valid representations for trajectory generation. The formula $\mathcal{L}_{KL}$ is as follows:

$$\mathcal{L}_{KL} = D_{KL}\left(q_\phi(z|Y,X) \| p_\phi(z|X)\right) = \int q_\phi(z|Y,X) \log \frac{q_\phi(z|Y,X)}{p_\phi(z|X)} dz \tag{20}$$

Since we assume $z$ follows a Gaussian distribution $\mathcal{N}(\mu, \sigma^2)$, and the prior $p_\phi(z|X)$ to be a standard normal distribution $\mathcal{N}(0, I)$, **Equation 20** is rewritten as:

$$\mathcal{L}_{KL} = \frac{1}{2} \sum_{j=1}^{dz} \left(\mu_j^2 + \sigma_j^2 - 1 - \log \sigma_j^2\right) \tag{21}$$

The third component is the interaction prediction loss $\mathcal{L}_{Int}$. It measures the binary cross-entropy between the predicted interaction behavior probability $\hat{B}$ and the ground truth label $B$ at each future time step. This loss is computed as:

$$\mathcal{L}_{Int} = -\frac{1}{M} \sum_{i=1}^{M} \sum_{t=T_{obs}+1}^{T_{obs}+T_{pre}} \left[-B_i^t \ln \hat{B}_i^t - \left(1-B_i^t\right)\ln\left(1-\hat{B}_i^t\right)\right] \tag{22}$$

To summarize, we combine all the losses into a weighted objective:

$$\mathcal{L}_{Fin} = w_1 \mathcal{L}_p + w_2 \mathcal{L}_{KL} + w_2 \mathcal{L}_{Int} \tag{23}$$

where $w_1$, $w_2$ and $w_3$ are hyperparameters.

## 5. Experiments

### 5.1 Experimental data

All post-crash LC samples are randomly divided into a training dataset (70%) and a test dataset (30%). To construct training and test samples, we apply a sliding window of length $T_{obs}+T_{pre}$ along each LC sample, with a step size of 0.5 s. For each resulting segment, we extract the historical trajectory $X$ and its graph $\mathcal{G}$ over $[1, T_{obs}]$ as input. The ground truth future trajectory $Y$ and the interaction behavior labels $B$ over the interval $[T_{obs}+1, T_{obs}+T_{pre}]$ are also extracted as outputs for model training and evaluation.

## 5.2 Baseline models

We consider the following models as baselines for comparison. To ensure a fair comparison, the data used to train our model are also used to train these baseline models. All these baseline models incorporate historical behavior, using either the historical trajectory $X$ or the graph data $\mathcal{G}$ as input. The output of each model is the predicted trajectory of the lane changer over the future horizon $[T_{obs}+1, T_{obs}+T_{pre}]$.

**(i) Convolutional Social-LSTM (CS-LSTM)** (Deo and Trivedi, 2018): This model extends the original Social-LSTM by incorporating a convolutional social pooling layer. In the encoder, historical trajectories $X$ are processed by an LSTM into hidden states. These hidden states are then arranged on a spatial grid to facilitate learning of social interactions via convolutional pooling. The decoder generates future trajectories using both the vehicles' motion history and the aggregated social context.

**(ii) Generative Adversarial Imitation Learning GRU (GAIL_GRU)** (Kuefler et al., 2017): This model adopts an adversarial imitation framework to predict future trajectories. Historical trajectories $X$ are passed to a GRU to generate hidden representations. A generator network uses these hidden states to produce the predicted trajectory. A discriminator network is used to distinguish between the generated trajectory and the ground truth trajectory.

**(iii) Social GAN (S-GAN)** (Gupta et al., 2018): This model employs a generative adversarial network (GAN) framework with an additional social pooling layer. The social pooling layer aggregates historical information from surrounding vehicles into a joint representation. Then, a generator and a discriminator are used to generate the future trajectory and to distinguish them from the real trajectory, respectively.

**(iv) GAT** (Diehl et al., 2019): This model takes graph $\mathcal{G}$ as input. A multi-head attention mechanism updates features of each node by attending over its neighbors. The updated node features are then decoded through a feed-forward network to predict future trajectory.

**(v) Spatio-Temporal Dynamic Attention Network (STDAN)** (Chen et al., 2022): This model applies a dynamic attention mechanism to consider both the spatial and temporal dependencies. For each observation time step, temporal attention captures the relevance of past states, while spatial attention identifies the most influential surrounding vehicles. The combined attention outputs are then passed to a decoder to predict future trajectory.

## 5.3 Evaluation Metrics

Consistent with prior works, we employ Average Displacement Error (ADE) and Final Displacement Error (FDE) to assess the trajectory prediction performance (Li et al., 2024; Zhou et al., 2023). ADE is defined as the average L2 distance between the predicted trajectory and the ground truth trajectory over the predicted horizon. FDE measures the L2 distance between the predicted trajectory and the ground truth trajectory at the final prediction time step. The formal definitions of ADE and FDE are given as follows:

$$ADE = \frac{1}{T_{pre}} \sum_{t=1}^{T_{pre}} \left\| \hat{Y}^{T_{obs}+t} - Y^{T_{obs}+T} \right\|_2 \tag{24}$$

$$FDE = \left\| \hat{Y}^{T_{obs}+T_{pre}} - Y^{T_{obs}+T_{pre}} \right\|_2 \tag{25}$$

For deterministic inference, ADE and FDE are computed directly for each test LC sample. For stochastic inference, we use the average ADE and FDE over $k$ predictions ($k = 20$) to assess model performance. Unlike some existing studies that report the minimum error across multiple samples, we adopt the average error to better reflect practical deployment scenarios. In real-world trajectory prediction, the ground-truth future is unknown at inference time, and it is therefore impossible to identify or select the trajectory with the minimum error. Reporting the minimum error may overestimate model performance by allowing occasional "lucky" samples, whereas the average error provides a more realistic and reliable assessment of predictive quality. Considering the short duration of LC maneuvers and driver reaction time, all experiments use an observation horizon of 1 s (10 time steps). To test the model's performance on short-term and long-term prediction, we set the prediction horizons to 1 s, 2 s, 3 s, 4 s, and 5 s.

## 6 Results

### 6.1 Ablation study

In this section, we conduct a series of ablation experiments to assess the contribution of the interaction-aware module to the post-crash LC trajectory prediction. Three base variants are considered: (i) a standalone CVAE; (ii) a pure Transformer-based predictor; and (iii) a CVAE with its decoder implemented as a Transformer network (CVAE+T). These variants are trained on the training dataset and tested on the test dataset. Their prediction performance on the test dataset, along with the results of our model, is summarized in **Table 1**.

Table 1 Ablation experiments of different components

| Prediction horizon | 1s | | 2s | | 3s | | 4s | | 5s | |
|---|---|---|---|---|---|---|---|---|---|---|
| | ADE | FDE | ADE | FDE | ADE | FDE | ADE | FDE | ADE | FDE |
| CVAE | 0.031 | 0.062 | 0.122 | 0.35 | 0.29 | 0.869 | 0.565 | 1.683 | 0.860 | 2.384 |
| Transformer | 0.025 | 0.058 | 0.125 | 0.347 | 0.302 | 0.851 | 0.540 | 1.502 | 0.892 | 2.238 |
| CVAE+T | 0.021 | 0.046 | 0.116 | 0.326 | 0.284 | 0.768 | 0.512 | 1.472 | 0.842 | 2.149 |
| CIT (ours) | 0.017 | 0.041 | 0.096 | 0.292 | 0.246 | 0.664 | 0.442 | 1.235 | 0.713 | 1.897 |
| Improvement ratio | **19.0%** | **10.9%** | **17.2%** | **10.4%** | **13.4%** | **13.5%** | **13.7%** | **16.1%** | **15.3%** | **11.7%** |

As we can see from **Table 1**, the standalone CVAE and Transformer frameworks exhibit comparable predictive performance. Replacing the CVAE decoder with a Transformer leads to moderate accuracy improvements across prediction horizons. Notably, incorporating the interaction-aware module into the CVAE+T model yields consistent performance gains across all horizons. Specifically, the proposed model reduces ADE by approximately 13%–19% and FDE by 10%–16%, compared with CVAE+T. These improvements are particularly evident in long-term prediction horizons, confirming the effectiveness of explicitly modeling interactions between the lane changer and the new follower. This module provides critical behavioral context and significantly enhances trajectory prediction accuracy.

### 6.2 Comparison Results

In this section, we compare the performance of our proposed model with the baseline models listed in **Section 5.2**. The comparison results are presented in **Table 2**. From **Table 2**, we observe that CS-LSTM achieves the lowest ADE and FDE among the baseline models. However, our proposed model consistently outperforms CS-LSTM across all prediction horizons. Specifically, even in short-term prediction horizons (1 s and 2 s), our model attains ADEs of 0.017 m and 0.096 m, improving upon CS-LSTM by 5.6% and 5.9%, respectively. In long-term prediction horizons, baseline models show a marked increase in ADE. For example, CS-LSTM reaches an ADE value of 0.803 m at a 5 s prediction horizon. Under the guidance of the interaction-aware module, our model reduces this to 0.713 m, representing an 11.2% improvement. In terms of FDE, CS-LSTM exhibits large errors of 1.437 m at a 4 s prediction horizon and 2.204 m at a 5 s prediction horizon. In contrast, our model achieves a reduction of 14.1% and 13.9% at the same horizons. These results demonstrate that our proposed model consistently delivers superior performance in post-crash LC trajectory prediction, both in short-term and long-term predictions.

Table 2 Comparison results

| Prediction horizon | 1s | | 2s | | 3s | | 4s | | 5s | |
|---|---|---|---|---|---|---|---|---|---|---|
| | ADE | FDE | ADE | FDE | ADE | FDE | ADE | FDE | ADE | FDE |
| S-GAN | 0.019 | 0.042 | 0.112 | 0.294 | 0.274 | 0.826 | 0.557 | 1.552 | 0.893 | 2.304 |
| CS-LSTM | 0.018 | 0.045 | 0.102 | 0.321 | 0.272 | 0.754 | 0.503 | 1.437 | 0.803 | 2.204 |
| STDAN | 0.082 | 0.087 | 0.157 | 0.370 | 0.352 | 0.937 | 0.641 | 1.683 | 0.912 | 2.461 |
| GAT | 0.096 | 0.157 | 0.151 | 0.383 | 0.342 | 0.923 | 0.756 | 1.891 | 0.917 | 2.439 |
| GAIL_GRU | 0.189 | 0.214 | 0.224 | 0.484 | 0.441 | 1.056 | 0.904 | 2.233 | 1.009 | 2.578 |
| **CIT (ours)** | 0.017 | 0.041 | 0.096 | 0.292 | 0.246 | 0.664 | 0.442 | 1.235 | 0.713 | 1.897 |
| Improvement ratio | **5.6%** | **8.9%** | **5.9%** | **9.0%** | **9.6%** | **11.9%** | **12.1%** | **14.1%** | **11.2%** | **13.9%** |

## 6.3 Qualitative analysis

In this section, we present several qualitative examples to visually illustrate the prediction performance of our proposed model. For comparison, we also include the prediction results from CVAE+T (our model without an interaction-aware module) and CS-LSTM (the strongest baseline). As shown in Table 2, our model achieves consistent improvements over CS-LSTM across all prediction horizons, with gains of approximately 5–9% in short-term scenarios and around 10–15% in long-term scenarios. To further illustrate the advantages of our proposed model, we focus on two long-term scenarios, namely 3 s and 5 s. For each horizon, we present two representative cases: one where the new follower exhibits non-yielding behavior and one where it yields.

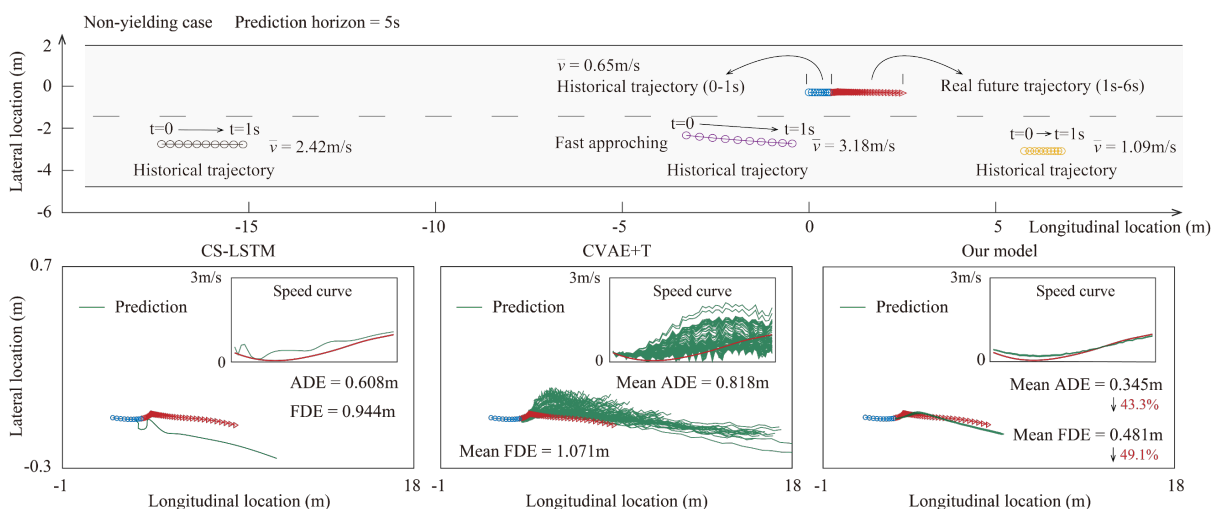

Figure 7 Predicted trajectory under a 5 s horizon with non-yielding behavior from the new follower

**Figure 7** depicts the 5 s prediction horizon scenario with non-yielding behavior. The upper part of the figure shows the historical movement of four vehicles: the lane changer (blue circles), the new leader (yellow circles), the new follower (brown circles), and the follower after the new follower (black circles). The red circles denote the lane changer's true trajectory over the next 5 s. From this figure, we can see that the lane changer moves slowly in the original lane, with an average speed of only 0.65 m/s. In contrast, the new follower moves at a higher speed of 3.18 m/s. This large speed difference prevents the lane changer from using this gap to complete the subsequent LC maneuver. As a result, the lane changer continues advancing slowly along the original lane over the next 5 s, awaiting a new opportunity to merge.

The lower panel of **Figure 7** compares the predicted lane changer's trajectory over the next 5 s using CS-LSTM, CVAE+T, and our proposed model. We can find that CS-LSTM produces a clearly implausible trajectory, with large lateral errors relative to the ground truth. Since both CVAE+T and our model use stochastic sampling, we drew 100 prediction samples to assess the variability. From the prediction results of CVAE+T, we notice that the predictions exhibit high variability across samples and significant deviation from the real trajectory. Although some predictions by CVAE+T align with the observed trajectory, most exhibit substantial divergence. The derived speed profiles also span unreasonable ranges. The average ADE and FDE of CVAE+T are 0.818 m and 1.071 m, respectively, both higher than those of CS-LSTM (ADE: 0.608 m, FDE 0.944 m). After incorporating the interaction-aware module, our model's predictions over 100 samples yield a mean ADE of 0.345 m and a mean FDE of 0.481 m, corresponding to error reductions of 43.3% and 49.1% relative to CS-LSTM, respectively. Visually, our model's predictions maintain controlled variability while discarding implausible trajectories, resulting in a close alignment with the observed lane changer's trajectory. The predicted speed profile also shows strong agreement with the true speed curve.

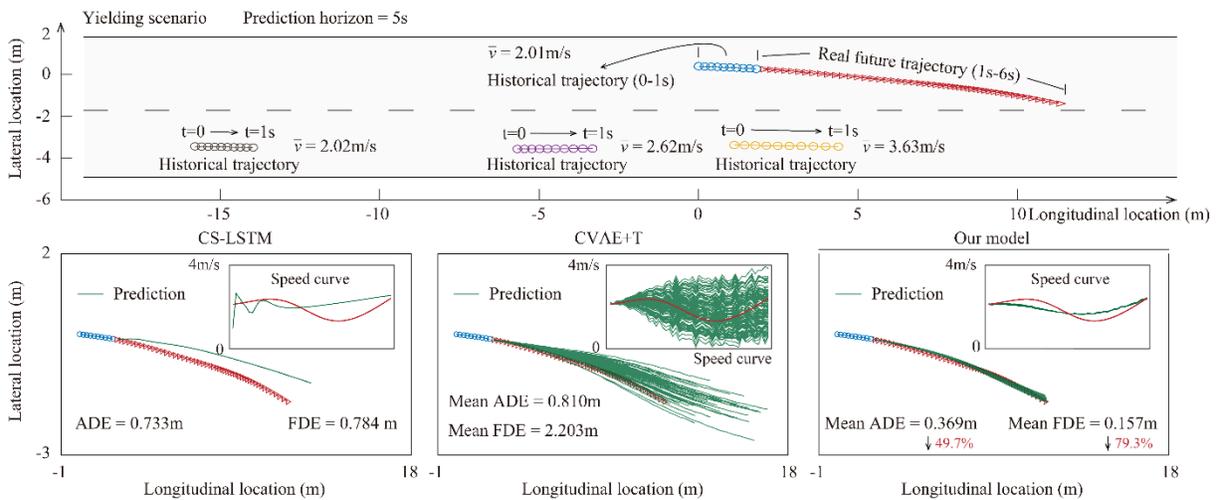

Figure 8 Predicted trajectory under a 5 s horizon with yielding behavior from the new follower

**Figure 8** shows the 5 s prediction horizon scenario in which the new follower yields to the lane changer. In this case, the lane changer moves at 2.01 m/s, which is much faster than that in **Figure 7**. In addition, the speed difference between the lane changer and the new follower is small. The new leader travels at 3.63 m/s, the highest speed among all other vehicles. This favorable gap condition allows the lane changer to complete the LC maneuver. Over the subsequent 5 s, the lane changer rapidly merges into the target lane, as shown by the red trajectory in the upper panel of **Figure 8**.

Prediction results for this scenario are given in the lower panel of **Figure 8**. For CS-LSTM, the predicted trajectory captures the general trajectory trend, with ADE and FDE values of 0.733 m and 0.784 m, respectively. The CVAE+T model still suffers from excessive variability, similar to the results observed in **Figure 7**. As observed in the results of CVAE+T in **Figure 7** and **Figure 8**, although the model can generate a variety of plausible lane changer trajectories via its latent space, the large fluctuations across samples compromise its practical prediction accuracy. The interaction-aware module in our model again filters out unrealistic predictions. Consequently, it achieves superior performance, with ADE decreasing from 0.733 m to 0.369 m and FDE decreasing from 0.784 m to 0.157 m.

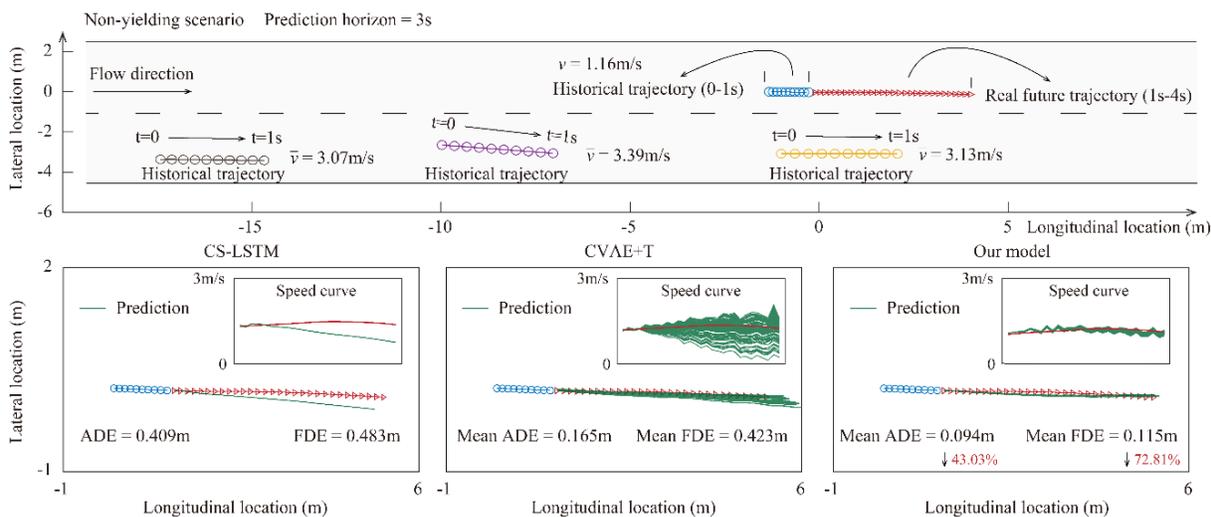
Figure 9 Predicted trajectory under a 3 s horizon with non-yielding behavior from the new follower

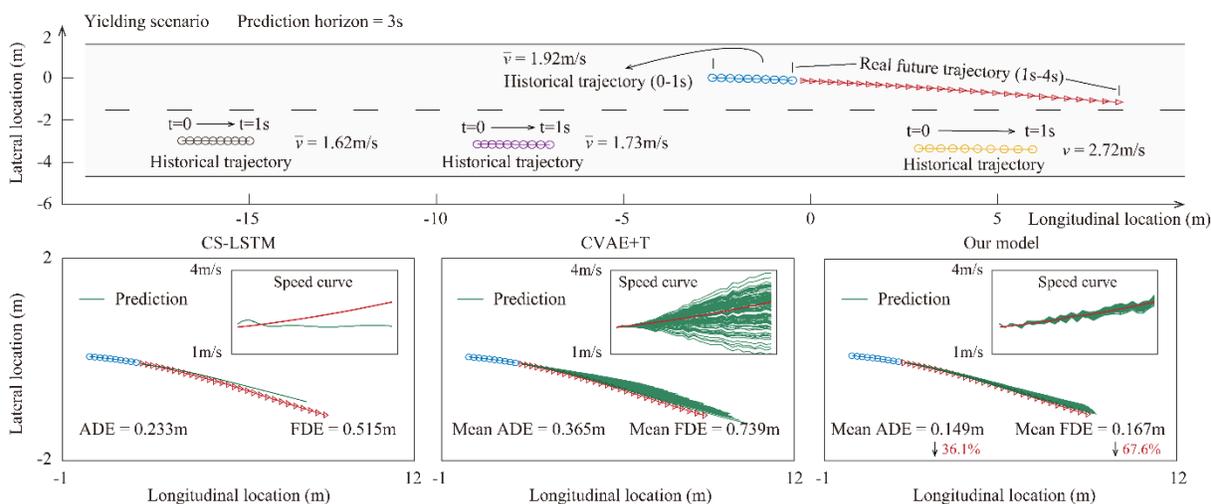
Figure 10 Predicted trajectory under a 3 s horizon with yielding behavior from the new follower

**Figure 9** and **Figure 10** present the non-yielding and yielding scenarios under a 3 s prediction horizon, respectively. As we can see from these figures, both CS-LSTM and CVAE+T produce lower errors compared to the 5 s prediction horizon. This reduction in error stems from the shorter prediction horizon, which reduces uncertainty and enhances model performance. This finding is consistent with the results in **Table 2**. Additionally, the prediction variability of CVAE+T is more concentrated around the real trajectories. Even so, our model with the interaction-aware module outperforms both CS-LSTM and CVAE+T by a clear margin. For the case in **Figure 9**, it reduces ADE from 0.165 m to 0.094 m (43.0% improvement) and FDE from 0.423 m to 0.115 m (72.8% improvement). For the case in **Figure 10**, ADE decreases from 0.233 m to 0.149 m (36.0% improvement) and FDE decreases from 0.515 m to 0.167 m (67.6% improvement). Furthermore, our model better controls prediction variability and generates speed curves that closely align with the observed speed profile. Overall, these examples visually demonstrate that our proposed model provides more accurate and stable post-crash LC trajectory predictions across varying scenarios.

**6.4 Crash risk test in trajectory prediction**

As discussed in **Section 3.2**, post-crash LCs are associated with a high risk of crashes. In such scenarios, small fluctuations in the predicted lane changer's trajectory may lead to inaccurate traffic safety analysis, for example, by overestimating crash risk and the risk of secondary crashes. Here, we aim to assess the crash risk implied by the predicted lane changer's trajectory using two indicators: false crash rate and the distribution of TTC. Since no real crash occurred in the post-crash LC samples in our test dataset, we consider a predicted lane changer's trajectory to be a false crash if it intersects with the true future trajectories of surrounding vehicles.

We compare the false crash rates of three models (CS-LSTM, CVAE+T, and CIT) across different prediction horizons using the test dataset. Given that CVAE+T and CIT models rely on stochastic inference, we conduct 20 inference runs per post-crash LC sample. If at least one of the 20 runs results in a crash, the sample is counted as a false crash case. The results are presented in **Figure 11(a)**. As shown in **Figure 11(a)**, CVAE+T exhibits a relatively high false crash rate across all prediction horizons. This is because its predicted trajectories are highly dispersed, increasing the likelihood of crashes. In contrast, our model consistently achieves a lower false crash rate than both CS-LSTM and CVAE+T across all prediction horizons. For example, at a 5 s horizon, our model's false crash rate is 1.9%, representing a 38.7% reduction compared to CS-LSTM (3.1%) and a 62.7% reduction compared to CVAE+T (5.1%). These results demonstrate that our model effectively reduces the false crash rate caused by trajectory prediction errors.

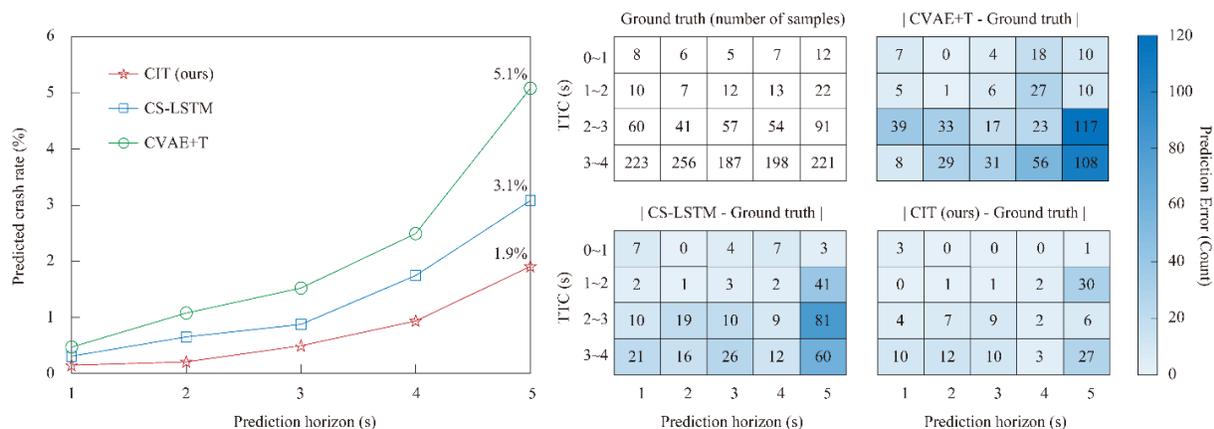

Figure 11 Crash risk test results: (a) false crash rate; (b) absolute deviation of TTC distributions from ground truth

In addition to the crash rate analysis, we also computed the two-dimensional TTC for post-crash LCs in the test dataset. The top-left panel of **Figure 11(b)** shows the ground truth, which represents the number of post-crash LC conflict samples falling within different TTC threshold ranges. We then calculate the TTC values between the predicted trajectories of the lane changer and surrounding vehicles using the three models. To highlight the differences from the ground truth, we report the absolute deviation between the predicted and ground truth counts for each TTC threshold range and prediction horizon. The results are given in **Figure 11(b)**. As shown, our model achieves the smallest deviations across all TTC thresholds and prediction horizons, followed by CS-LSTM and CVAE+T. These results further indicate the superior performance of our model in analyzing potential post-crash LC conflicts.

**6.5 Transferability test**

To evaluate the model's transferability, we construct two additional post-crash LC datasets. The first dataset is collected on an expressway referred to as the Inner Ring East (site 2) in Nanjing. The road structure of site 2 is illustrated in the left part of **Figure 12**. Drone videos at this location are recorded during the morning peak hours (7:30-9:30 AM) from September 24, 2023, to August 1, 2024. To further evaluate the model's performance in different cities, we also collect post-crash LC samples in Chongqing, China, at a merging section on Haixia Road, referred to as site 3 (right part of **Figure 12**). The recording period at site 3 spanned from March 19, 2024, to November 26, 2024, during the morning peak hours (7:00-9:00 AM) each sunny workday. Vehicle trajectory data are extracted from all collected drone videos. Only post-crash LC samples are retained for analysis, as illustrated in the lower part of **Figure 12**. In total, 325 post-crash LC samples are obtained at site 2, and 117 such samples are extracted at site 3.

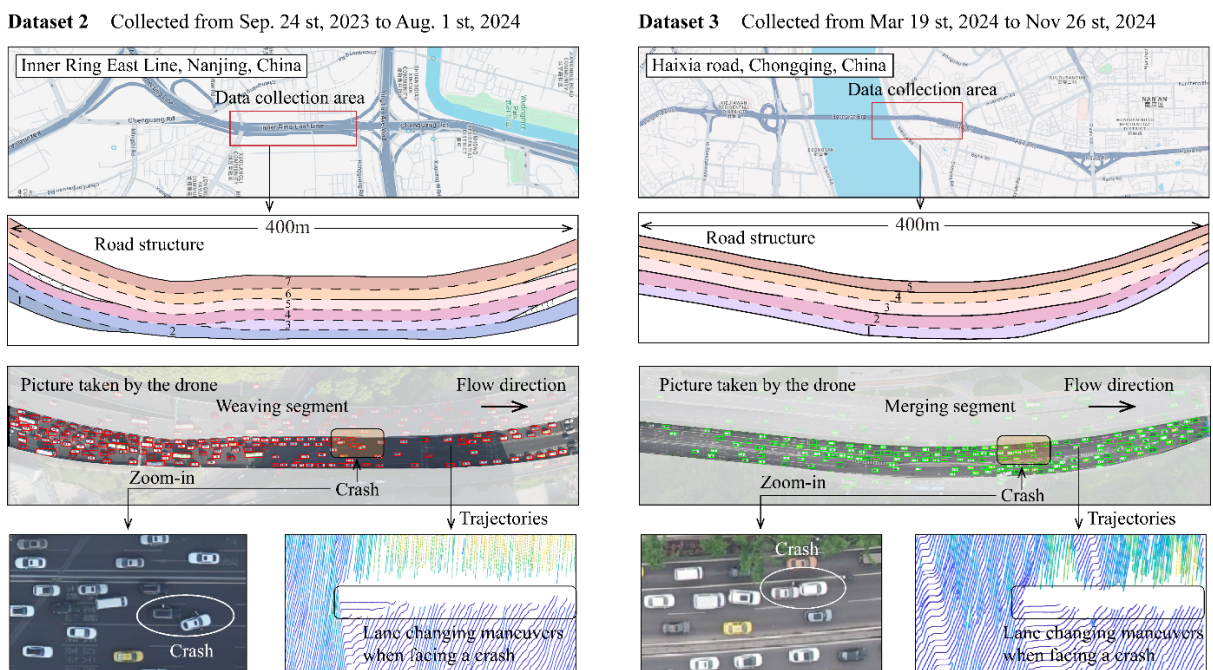

Figure 12 Two additional study sites for transferability test

Following the same sliding-window procedure described in **Section 5.1**, we generate prediction segments from post-crash LC samples at site 2 and site 3. We then directly apply the trained CIT model along with the strongest baseline among the compared methods (CS-LSTM) to predict the trajectories of the lane changers in these segments. Prediction performance is assessed using ADE and FDE. The evaluation results for site 2 and site 3 are summarized in **Table 3** and **Table 4**, respectively.

Table 3 Transferability test results for study site 2

| Prediction horizon | 1s | | 2s | | 3s | | 4s | | 5s | |
|---|---|---|---|---|---|---|---|---|---|---|
| | ADE | FDE | ADE | FDE | ADE | FDE | ADE | FDE | ADE | FDE |
| CS-LSTM | 0.019 ↑5.6% | 0.049 ↑8.9% | 0.113 ↑10.4% | 0.355 ↑10.6% | 0.308 ↑13.4% | 0.833 ↑10.5% | 0.567 ↑12.7% | 1.591 ↑10.7% | 0.917 ↑14.2% | 2.427 ↑10.1% |
| CIT (ours) | 0.018 ↓6.3% | 0.042 ↑2.7% | 0.102 ↑6.6% | 0.306 ↑4.8% | 0.251 ↑4.0% | 0.710 ↑6.9% | 0.470 ↑6.3% | 1.328 ↑7.5% | 0.761 ↑6.7% | 1.994 ↑5.1% |

Table 4 Transferability test results for study site 3

| Prediction horizon | 1s | | 2s | | 3s | | 4s | | 5s | |
|---|---|---|---|---|---|---|---|---|---|---|
| | ADE | FDE | ADE | FDE | ADE | FDE | ADE | FDE | ADE | FDE |
| CS-LSTM | 0.021 ↑16.7% | 0.051 ↑13.3% | 0.114 ↑12.1% | 0.356 ↑10.9% | 0.324 ↑19.3% | 0.861 ↑14.2% | 0.585 ↑16.3% | 1.612 ↑12.2% | 0.970 ↑20.8% | 2.491 ↑13.0% |
| CIT (ours) | 0.018 ↑6.5% | 0.044 ↑8.1% | 0.105 ↑9.7% | 0.301 ↑6.3% | 0.267 ↑8.4% | 0.722 ↑8.7% | 0.480 ↑8.5% | 1.356 ↑9.8% | 0.784 ↑9.9% | 2.003 ↑5.6% |

As shown in **Table 3**, the performance of CS-LSTM degrades noticeably when directly applied to site 2. An exception is observed at the 1 s prediction horizon, where the performance degradation remains within 10%. In other settings, the performance decline exceeds 10%, with the most notable case at the 5 s prediction horizon, where the ADE increases from 0.803 m to 0.917 m (a 14.2% increase). In contrast, our proposed model exhibits more stable performance across all prediction horizons, with the increase in prediction error remaining between 2% and 8%. As illustrated in **Table 4**, the performance of CS-LSTM deteriorates further when transferred to a different city (site 3 in Chongqing). Even at the 1 s prediction horizon, the ADE increases by 16.7%, and at 5 s, the degradation reaches 20.8%. While our proposed model also experiences a decline at site 3, the ADE increase remains below 10%. Given the more complex driving environments at site 2 and the cross-city transfer to site 3, the slight increase in error exhibited by our model remains within an acceptable range. Compared with CS-LSTM, these results indicate that our model generalizes well across different sites with only moderate performance variation, indicating strong transferability.

## 7. Conclusion

In this paper, we investigate post-crash LC maneuvers with a focus on behavioral features and trajectory modeling. A one-year drone video recording at a merging section in Nanjing, China was conducted to collect post-crash LC samples. A total of 1,374 post-crash LC samples were obtained for analysis and model development. Several key behavioral features of post-crash LCs are examined, including the yielding behavior of the new follower, duration of LC, insertion speed, speed difference between the lane changer and the new follower, and crash risk measured by two-dimensional TTC. These features are also compared against those of MLCs and DLCs.

Among all differences, the yielding behavior of the new follower emerges as the most distinctive feature of post-crash LCs. We find that 79.4% of post-crash LCs experience at least one instance of non-yielding behavior by the new follower before the LC maneuver is completed, compared to only 21.7% for DLCs and 28.6% for MLCs. This non-yielding behavior substantially increases the complexity of post-crash LC maneuvers. In such situations, lane changers have to remain alert to the driving behavior of successive new followers and adapt their movement in response to the availability of gaps. This increased complexity is mainly due to the large speed difference between the lane changer and the vehicles in the target lane. We further find that the duration of non-yielding behavior accounts for nearly 50% of the entire LC maneuver time in post-crash LCs. Moreover, 41% of post-crash LCs involve a TTC value below 2 s, indicating a high level of crash risk. These empirical results indicate that post-crash LCs are distinct from normal LCs and deserve more attention.

Moreover, we propose a deep learning framework for post-crash LC trajectory prediction by explicitly incorporating the yielding behavior of the new follower. A CVAE is developed to capture the uncertainty in driving behavior. A graph-based network is introduced to model the interaction between the lane changer and the new follower. Its output is used to guide the latent space of the CVAE. A Transformer-based decoder is then applied to generate the future trajectory of post-crash LCs.

Ablation experiments confirm that the interaction-aware module significantly improves model accuracy. Compared with the CVAE+T model, it reduces ADE by 13.4%–19.0% and FDE by 10.4%–16.1% across prediction horizons from 1 s to 5 s. These results demonstrate that interaction behaviors among vehicles can effectively guide deep learning models for trajectory prediction. Further comparisons with five baseline models demonstrate that our proposed model achieves superior performance across all horizons, particularly in long-term prediction. In addition, the crash risk test demonstrates that the predicted trajectories from our model provide better interpretability for traffic safety analysis. Finally, we validate the model's transferability using post-crash LC datasets collected from different sites and cities.

Although this study is the first attempt to empirically analyze and model post-crash LC maneuvers, several limitations remain. First, more studies can be conducted to explore other aspects of post-crash LCs, including the decision-making process and their impact on traffic operations. Second, our trajectory prediction framework targets only the trajectory of the lane changer. Future studies should consider simultaneous trajectory prediction of surrounding vehicles to better support autonomous driving and traffic management. Finally, an important future direction is to develop motion planning approaches for autonomous vehicles that can handle post-crash LC scenarios in mixed traffic flow.

## ACKNOWLEDGMENTS

This work was supported by the National Natural Science Foundation of China (52525204, 52232012). The authors thank the anonymous reviewers for their time to review our article and their constructive comments.